\documentclass[final]{cvpr}

\usepackage{times}
\usepackage{epsfig}
\usepackage{graphicx}
\usepackage{amsmath}
\usepackage{amssymb}

\usepackage[pagebackref=true,breaklinks=true,colorlinks,bookmarks=false]{hyperref}

\def\cvprPaperID{11597} 
\def\confYear{CVPR 2021}

\begin{document}

\title{Unsupervised Shadow Removal Using Target-Consistency Generative Adversarial Network}

\author{Chao Tan\\
School of Computer Science and Engineering\\
Chongqing University of Technology\\
{\tt\small istvartan@outlook.com}
\and
Xin Feng\\
School of Computer Science and Engineering\\
Chongqing University of Technology\\
{\tt\small xfeng@cqut.edu.cn}
}

\maketitle

\begin{abstract}
  In this paper, we develop a simple yet effective target-consistency generative adversarial network (TC-GAN) for the shadow removal task in an unsupervised manner. Compared with the bidirectional mapping in cycle-consistency GAN based methods for unsupervised shadow removal, TC-GAN targets to learn a unidirectional mapping to translate shadow images into shadow-free ones. With the proposed target-consistency constraint that is designed to connect two dual GAN-based sub-networks, the correlations between shadow images and the output shadow-free image, and the realness of recovered shadow-free image are strictly confined. Extensive comparison experiments results show that TC-GAN outperforms the state-of-the-art unsupervised shadow removal methods by 14.9\% in terms of FID and 31.5\% in terms of KID. It is rather remarkable that TC-GAN achieves comparable performance with supervised shadow removal methods.
\end{abstract}

\section{Introduction}

To recognize and remove shadows from the natural images is very important and critical for many computer vision tasks such as the object detection, recognition and tracking, etc \cite{phan2020improved,long2014accurate,luo2019end,hua2017collaborative,long2015multi}. In principle, shadow removal can be formulated as an image restoration problem, where a shadow-free image $I_Y$ is either recovered from the element-wise addition of input shadow image $I_X$ and a negative residual that is generated from the shadowed image (denoted by $G_{res}(I_X)$ in equation \ref{eq1}) \cite{zhang19,ding19,tang2020sdrnet}, or the linear transformation of $I_X$ and a shadow matte (or shadow factors) \cite{le2019shadow,qu17,amin2020automatic}.
\vspace{-0.5pt}
\begin{equation}\label{eq1}
I_Y = I_X \oplus G_{res}(I_X)
\end{equation}

In practice, because of environmental uncertainties, such as different depth of shadows projected by different material of objects and illumination conditions, similar visual features between shadow regions and some objects in the image, etc., the task of shadow removal is quite challenging and has always been a fundamental research topic. 

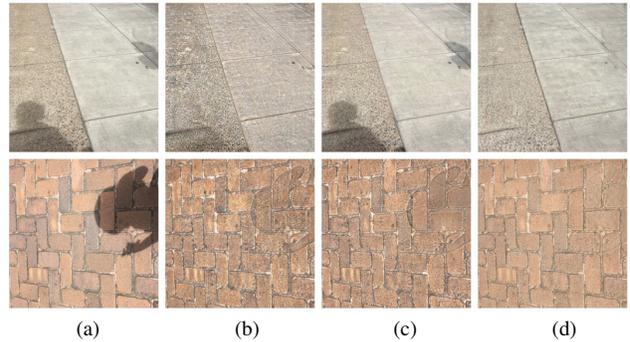
\begin{figure}[t]
	\begin{center}
		\includegraphics[width=1.0\linewidth]{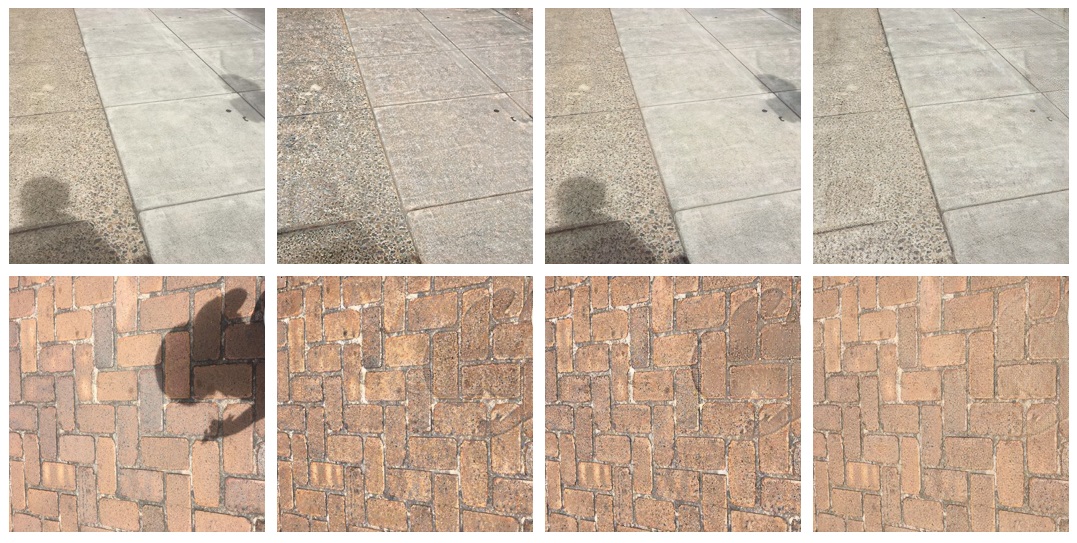}
		\put(-210,-8){\footnotesize (a)}
	    \put(-150,-8){\footnotesize (b)}
	    \put(-90,-8){\footnotesize (c)}
	    \put(-30,-8){\footnotesize (d)}
	\end{center}
	\vspace{-0.2pt}
	\caption{Visualization of the shadow removal result. (a) is input shadow images, (b) is results of CycleGAN \cite{zhu17}, (c) is results of Mask-ShadowGAN \cite{hu19} and (d) is results of proposed TC-GAN.}
	\label{fig:fig1}
\end{figure} 
\vspace{-0.5pt}

Traditional shadow removal methods are usually developed to localize and remove shadows by  designing hand-crafted color and illumination features based on some prior assumption, e.g.the consistent illumination in shadow region \cite{Gryka15,Guo12,vicente17,yang12}. Recently, deep learning based shadow removal approaches have achieved remarkable performance \cite{ding19,hu18,qu17,wang18}. In these methods, the shadow matte or mask and the removal operation are learned from the paired shadow removal datasets in the supervised manner. In particular, the generative adversarial learning \cite{goodfellow14}, which learns a non-linear mapping network through restricting the output shadow-free images to be indistinguishable from the non-shadow label images, have shown to be effective for the shadow removal task \cite{ding19,sidorov19,xiaodong19}. However, the problem of this kind of supervised learning is that high quality paired shadow and non-shadow images are difficult and costly to obtain in practice. Comparatively, the unpaired natural scene images that contain shadows and non-shadow images are easy to be acquired. This allows us to collect a large number of shadow images and mismatched non-shadow images with variety scenes for unsupervised shadow removal learning. However, without the corresponding shadow-free labels for supervision, the unsupervised shadow removal is much more difficult so that few existing works has achieved satisfactory results so far, as the results of CycleGAN \cite{zhu17} and Mask-ShadowGAN \cite{hu19} shown in Figure \ref{fig:fig1}.
\vspace{-0.5pt}
\begin{figure}[t]
	\begin{center}
		\includegraphics[width=1.0\linewidth]{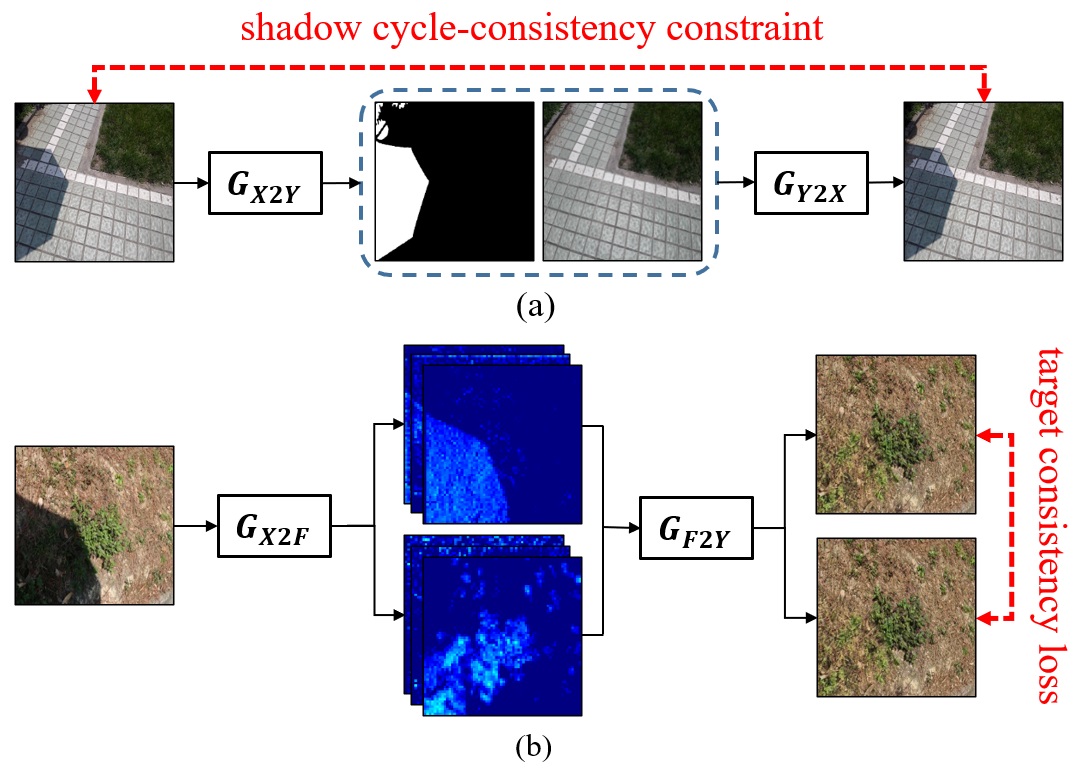}
	\end{center}
	\caption{Comparison of shadow cycle-consistency constraint in Mask-ShadowGAN \cite{hu19} (a) and target-consistency constraint in TC-GAN (b). The red dotted lines denote the unsupervised constraint.}
	\vspace{-0.5pt}
	\label{fig:fig2}
\end{figure}
\vspace{-2pt}

Inspired by the recent unsupervised shadow removal work \cite{hu19} that casts the unsupervised shadow removal task into image-to-image translation problem based on cycle-consistent adversarial network, in this work, we present a target-consistent generative adversarial network (TC-GAN) that aims at translating the input shadow image into shadow-free image with unpaired data in the unsupervised manner. Unlike the bidirectional mapping (denoted by $G_{X2Y}$ and $G_{Y2X}$ separately in Figure \ref{fig:fig2} (a)) in the cycle-consistency constraint-based models for unsupervised learning in \cite{hu19}, which requires the generated shadow mask to reconstruct the image from the non-shadow domain back to the shadow domain, TC-GAN learns a one-sided mapping network $G_{X2Y}$ to directly translate the shadow image into shadow-free image, as shown in Figure \ref{fig:fig2} (b). 

In particular, TC-GAN draws inspiration from two facts. One fact is that multiple shadow-discrepant images may correspond to the same shadow-free image. In other words, if the input of $G_{X2Y}$ has two (could have more) shadow-discrepant images (or features) that correspond to one shadow-free image, the output of $G_{X2Y}$ that contains two shadow-free images should be consistent. Here, we propose a target consistency constraint to supervise the content consistency between the two outputs of $G_{X2Y}$. Now the problem is how we have two shadow-discrepant images (or features) for a given input shadow image. In this work, we explicitly design a set of shadow transform encoders (STE) in the generative process of TC-GAN to transform the input shadow image into two feature representations (we denote as $G_{X2F}$ in Figure \ref{fig:fig2}), which learns to encode the different intrinsic compositions between the non-shadow features and shadow residuals. The other fact is that different non-shadow natural images shall share some common illumination and brightness characteristics, for example, generally without dramatic brightness variations. This motivates us to adapt the discriminator in TC-GAN to learn the hidden common features from the large-scale non-shadow natural image dataset, which is available in the unsupervised shadow removal dataset.

To realize the overall process, we design TC-GAN as a dual generative adversarial network structure. Through recognizing different real non-shadow images from the generated shadow-free images of the corresponding discriminator, each generative adversarial sub-network has different adversarial optimization objective. Along with target consistency constraint during training, the whole network guarantees that: (i) different feature embedding can be generated from the input shadow image, (ii) the decoded shadow-free images from two generators not only have consistent content but also agree with real shadow-free image. Finally, we have a simple model selection module to select the best output shadow-free result from the two generators.

The proposed unidirectional mapping generative adversarial network solves the unsupervised shadow removal problem from a new perspective. As extensive comparison experiments on the challenge unsupervised shadow removal dataset are shown, TC-GAN achieves the best performance in terms of both quantitative and qualitative comparison over state-of-the-art unsupervised shadow removal approaches. Furthermore, on the widely use supervised shadow removal dataset, TC-GAN even obtains comparable results with recent excellent supervised methods.

\vspace{-0.5pt}
\begin{figure*}[t]
	\begin{center}
		\includegraphics[width=1.0\linewidth]{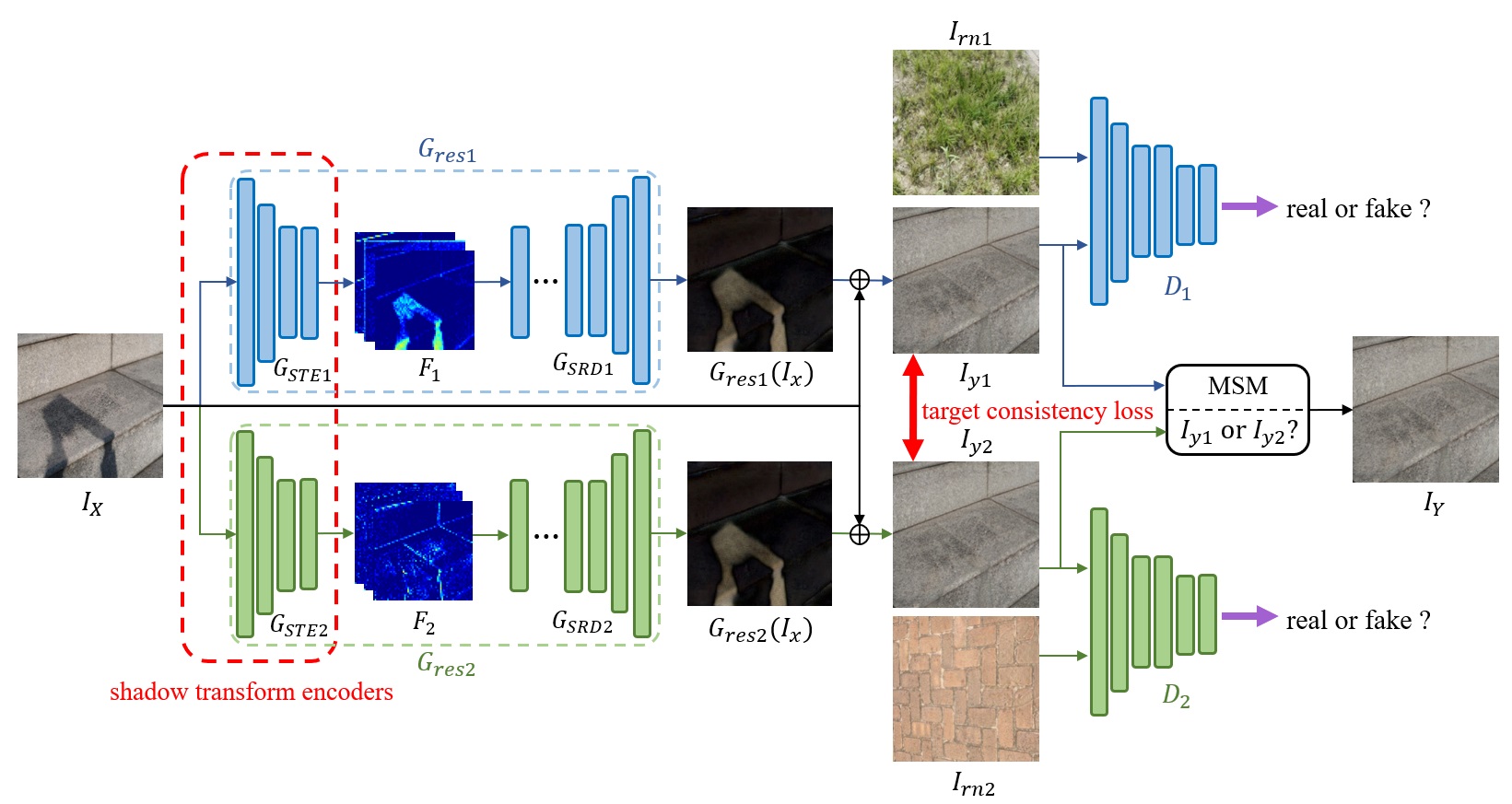}
	\end{center}
	\vspace{-0.2pt}
	\caption{An illustration of the proposed target-consistence generative adversarial network (TC-GAN).}
	\label{fig:fig3}
\end{figure*}
\section{Related Work}

\subsection{Shadow Removal}
Natural scene shadow removal has been widely studied. Early methods usually focused on physically modeling the shadow image and estimating the model parameters in terms of gradient domain manipulation \cite{finlayson05,liu08,mohan07}, shadow illumination and color transfer \cite{vicente17,khan15,reinhard01,xiao13}, and accurate shadow matte \cite{Gryka15,Guo12,wu07}. Recently, driven by shadow removal datasets with paired shadow and shadow-free images \cite{qu17,wang18}, deep learning-based methods are widely used to analyze the potential mapping relation between shadow images and corresponding non-shadow images \cite{ding19,sidorov19,Lin_2020_CVPR,wei19}. Qu \etal \cite{qu17} proposes an end-to-end multi-context embedding network to integrate high-level semantic context for shadow removal. Zhang \etal \cite{zhang19} uses a unified end-to-end framework to explore the residual and illumination for shadow removal. For learning shadow detection and removal jointly, Wang \etal \cite{wang18} proposes a stacked conditional generative adversarial network, and Hu \etal \cite{hu18} presents a deep neural network for shadow detection and removal by analyzing the spatial image context in a direction-aware manner.

Since now, a few of methods have explored the unsupervised shadow removal on the unpaired dataset and the only unsupervised dataset has been proposed in \cite{hu19}. In paticular, Hieu and Dimitris \cite{le2020shadow} use shadow and non-shadow patches cropped from the shadow images and train a physical shadow formulation-based model via an adversarial framework. Hu \etal \cite{hu19} present a Mask-ShadowGAN that learns shadow removal from the unpaired training data. The method takes use of the cycle consistency of CycleGAN \cite{zhu17}, and with the guidance of the generated binary mask in the generative process, it facilitates the bidirectional mapping to transform the shadow image into shadow-free image.

\subsection{Unsupervised Adversarial Learning}
Unsupervised shadow removal can be regarded as a domain mapping problem. Recent domain mapping works have used adversarial learning \cite{goodfellow14} to solve unpaired domain mapping \cite{johnson16,zhang18b,li18}. Generative adversarial networks \cite{goodfellow14} are powerful adversarial learning models, which learn a generator and a discriminator jointly such that the generator produces realistic images that confuse the discriminator \cite{arjovsky17,mao17,salimans16}. The cycle consistency constraint based methods \cite{zhu17,kim17,yi17} learn a bijection mapping by enforcing cycle consistency between the reconstructed data and input one to produce convincing mappings from the source domain to the target domain without paired supervision. Since then, many subsequent methods have been proposed to improve the cycle-consistency constraint \cite{anoosheh18,li18b,kim19,tang19a,yang19}. Furthermore, for one-sided unsupervised domain mapping, Benaim and Wolf \cite{benaim17} propose to maintain the distances between samples within domains, and Fu \etal \cite{fu19} develop a geometry-consistent generative adversarial network to perform image-to-image translation.
 

\section{TC-GAN for unsupervised Shadow Removal}
Without paired shadow and non-shadow images for supervision, the non-linear mapping $G_{X2Y}$ cannot directly learn from the unpaired dataset. \cite{hu19} takes use of CycleGAN structure to render a backward mapping $G_{Y2X}$ with the help of generated mask from $G_{X2Y}$, as shown in Figure \ref{fig:fig2} (a), and $I_X$ is used as the reference to supervise the reconstructed image. Unlike the CycleGAN based unsupervised shadow removal that dependents on the predicted shadow mask, TC-GAN is an unidirectional mapping network to directly learn $G_{X2Y}$ based on generative adversarial learning framework. Mainly motivated by the fact that if two different shadow images were captured by shading differently on one non-shadow image, the recovered shadow-free image $I_Y$ from the two different shadow images should be the same, we build TC-GAN as a dual generator-discriminator GAN structure to conform this assumption into network learning. The overall architecture of TC-GAN is shown in Figure \ref{fig:fig3}. 

\subsection{Generative Network in TC-GAN}

For the input shadow image $I_X$, we design two generators $G_1$ and $G_2$ to learn two separate $G_{X2Y}$ mappings, deriving two shadow-free images $I_{y1}$ and $I_{y2}$ with respect to $I_X$, respectively. Here, on the one hand, in order to confine $I_{y1}$ and $I_{y2}$ to be content consistent, we propose a target consistency constraint to act on the ends of the two generators. On the other hand, the two generators are acquired to produce two different representations of $I_X$ to mimic the assumption that two different shadow-discrepant images correspond to the same non-shadow image. 

The two generators are mainly composed of two encoder-decoders with dual network structure, which we term as shadow residual generators, with $G_{res1}$ and $G_{res2}$ indicating the related sub-network in Figure \ref{fig:fig3}.  The two sub-networks are expected to generate two different shadow residual images from $I_X$. In particular, the encoder in each of the generator is a shadow transform encoder (denoted by $G_{STE1}$ and $G_{STE_2}$ for each encoder in Figure \ref{fig:fig3}). During network training, $G_{STEi},(i \in \{1,2\})$ learns to transform $I_X$ from image domain to a kind of feature compositions, i.e. $F_i = G_{STEi}(I_X),i \in \{1,2\}$, which encodes the inherent relations between the non-shadow and shadow residual features with respect to $I_X$. The subsequent shadow residual decoders $G_{SRD1}$ and $G_{SRD2}$ are designed to decode the feature representations from feature domain to shadow residual images, i.e. $G_{resi}(I_X) = G_{SRDi}(F_i),\in \{1,2\})$, layer by layer. At last, by applying the element-wise addition between $I_X$ and $G_{res1}(I_X)$ and $G_{res2}(I_X)$, respectively, the two separate shadow-free images $I_{y1}$ and $I_{y2}$ can be recovered. 

In order to ensure $G_1$ and $G_3$ producing two separate shadow-free images from $I_X$, we intentionally initialize the dual sub-networks with different parameters, and the differences are maintained during the whole training process through different adversarial optimization objectives. Ideally, $I_{y1}$ and $I_{y2}$ will be exactly the same images under the target consistent constraint, however, as they were trained by different adversarial generators, we build a model selection module (MSM) to make a selection from the two shadow-free images. The MSM module is a pre-trained shadow/non-shadow binary classifier, which is embedded at the end of the generative network in TC-GAN and only applied for the network inference phrase. Through comparison of the output probability of the two classes, it selects the one with higher confidence from the two generated images ($I_{y1}$ and $I_{y2}$) to be a real non-shadow image as the final output.

To summarize, the overall process of the generative network of TC-GAN can be formulated as:

\begin{equation}\label{eq2}
\begin{aligned}
I_Y = \text{MSM}(I_X + G_{res1}(I_X)), \\
      I_X + G_{res2}(I_X)))
\end{aligned}
\end{equation}

In model implementation, we build the shadow transform encoders with three convolution layers (stride-2) and nine residual blocks \cite{yi17} , and the shadow residual decoders are designed to have three stride-2 deconvolution layers. Each layer in both encoders and decoders is followed by an instance normalization and ReLU activation function. The structure of MSM is built by four convolution layer with stride of 2, and each layer uses the same nomalization and activation function with encoders. 

\subsection{Discriminative Network in TC-GAN}
Similar with the generative network, there are two discriminators $D_1$ and $D_2$ corresponding to the two generators and with the same network structure in the Discriminative network. The discriminator basically contains four stride-2 convolutional layers, each of which is followed by an instance normalization and Leaky ReLU activation function (slope of 0.2).  

In the GAN based image-to-image translation approaches for unpaired data, image data in the same domain is believed to share some common characteristics. Delighted by the assumption and the fact that images from the non-shadow domain have agreement on some aspects like consistent illumination and brightness, etc., we facilitate each discriminator in the dual discriminative network for adversarial training by feeding a randomly selected sample from non-shadow image dataset as the real non-shadow image (as we denote by $I_{rn1}$ and $I_{rn2}$ for $D_1$ and $D_2$ in Figure \ref{fig:fig3}) for discrimnative supervision. In addition, to leverage the two generators for individual learning, we intentionally choose two different real non-shadow images from the dataset, i.e., $I_{rn1} \neq I_{rn2}$, for the two discriminators.

\subsection{Target Consistency Constraint}

As before mentioned, the target-consistency constraint is proposed to confine the correlation between $I_{y1}$ and $I_{y2}$. Specifically, given an input sample $x$ being drawn from the shadow image data distribution $P_X$, the output targets of the two generators $G_1$ and $G_2$ are restricted to be as close as possible by applying the target-consistence constraint:
\begin{equation}\label{eq3}
\begin{aligned}
\mathcal{L}_{tc}(G_1,G_2) = \mathcal{E}_{x \sim P_x}[\| (x + G_{res1}(x)) \\
- (x + G_{res2}(x)) \|_{1}]
\end{aligned}
\end{equation}

The key role of target consistency constraint is to connect the dual sub-GAN networks ( $G_1$-$D_1$ and $G_2$-$D_2$) for the overall learning of TC-GAN, but more than that, it drives an adversarial learning between the consistency of $I_{y1}$ and $I_{y2}$ and distinctiveness of $G_1$-$D_1$ and $G_2$-$D_2$, in addition to each adversarial constraint of $G_1$-$D_1$ and $G_2$-$D_2$. Furthermore, the target-consistence loss restricts the output shadow-free targets directly, whereas random disturbance would be induced by using the binary mask guided cycle-consistence constraint in \cite{hu19}, leading to ambiguous textures in the recovered shadow-free image as shown in Figure \ref{fig:fig1} (c).

\subsection{Fully Objective}
For training each sub-GAN network, we use the generic adversarial constraint \cite{goodfellow14} to enforce the generator and discriminator to jointly optimize the adversarial loss. Specifically, let $p(X)$ and $p(Y)$ represent the data distribution of shadow domain $D_X$ and non-shadow domain $D_Y$, respectively, the adversarial loss $\mathcal{L}_{gan}(G_i,D_i)$ for each sub-GAN $i \in {1,2}$ can be expressed as:
\begin{equation}\label{eq4}
\begin{aligned}
\mathcal{L}_{gan}(G_i,D_i) = \mathcal{E}_{y_i \sim P_Y}[logD_i(y_i)]\\
+\mathcal{E}_{x \sim P_X}[log(1-D_i(x+G_{resi}(x)))]
\end{aligned}
\end{equation}

where $y_1$ and $y_2$ are the two different real non-shadow images $I_{rn1}$ and $I_{rn2}$ for supervising true shadow-free image restoration.

In addition, by following most of the GAN-based image-to-image translation works on unpaired data \cite{taigman16}, we further use the identity loss to leverage the generators learning from real non-shadow images. For the purpose of (i) preserving the consistency between the generated shadow-free image and input image on the non-shadow areas in terms of color or other appearance features \cite{zhu17}. (ii) making faster convergence for training on both generators, the identity loss of $G_1$ and $G_2$ for non-shadow data $y_1$ and $y_2 \in P(Y)$ are described as:

\begin{equation}\label{eq5}
\begin{aligned}
\mathcal{L}_{idt}(G_i) = \mathcal{E}_{y_i \sim P_Y}[\| y_i + G_{resi}(y_i) - y_i\|_{1}]  \\
                      =\mathcal{E}_{y_i \sim P_Y}[\|G_{resi}(y_i)\|_{1}]
\end{aligned}
\end{equation}

In summary, the final loss function for TC-GAN ($\mathcal{L}_{tcgan}$) is a weighted sum of the adversarial loss of the both sub-GANs, target-consistency loss, and identity loss:
\begin{equation}\label{eq6}
\begin{aligned}
\mathcal{L}_{tcgan}(G_1, G_2, D_1, D_2) & = \\
\lambda_{1} \cdot (\mathcal{L}_{gan}(G_1,D_1) + \mathcal{L}_{gan}(G_2,D_2)) \\
+ \lambda_{2} \cdot \mathcal{L}_{tc}(G_1,G_2) \\
+ \lambda_{3} \cdot (\mathcal{L}_{idt}(G_1) + \mathcal{L}_{idt}(G_2)),
\end{aligned}
\end{equation}
where $\lambda_{1}$,$\lambda_{2}$ and $\lambda_{3}$ are trade-off hyperparameters to weight the contribution of $\mathcal{L}_{gan}$, $\mathcal{L}_{tc}$ and $\mathcal{L}_{idt}$, respectively. In this work, We empirically set $\lambda_{1}$,$\lambda_{2}$ and $\lambda_{3}$ as 1, 40, and 5 separately in all the experiments. Furthermore, the negative log likelihood objective is replaced by a least-square loss for adversarial loss, which is shown to be more stable and effective than the original log likelihood according to \cite{mao17}. At last, we optimize the whole network in a minimax optimizer.

\vspace{-2pt}
\subsection{Implementation Details}
In training process, we use the Adam solver \cite{kingma14} with a basic learning rate of 0.0002, and assign the first and second momentum values as 0.5 and 0.999 for network optimization. The parameters for all the generators and discriminators are random initialized by a zero-mean Gaussian distribution with a standard deviation of 0.02. The network is trained in a total of 200 epochs with each mini-batch size of one. In addition, we fix the learning rate for the first 100 epochs, and gradually reduce it to be zero by applying a linear decay rate for the rest of epochs. All the images are first re-scale to 286 ×286 and randomly cropped to be size of 256 ×256 as inputs. Lastly, we built our model on PyTorch library and trained on a computer with a single NVIDIA GTX1080Ti GPU.

\section{Experiments}
In this work, we mainly evaluate our proposed TC-GAN on the unsupervised shadow removal dataset. However, in order to further validate the method with annotated data and make more comparison, we perform training and experiment on the supervised shadow removal dataset as well.  
\subsection{Shadow Removal Datasets}
\textbf{The supervised dataset}: The ISTD \cite{wang18} is a widely used dataset for supervised shadow removal. It contains 1330 training triplets and 540 testing triplets of shadow image, non-shadow image and binary shadow mask. The shadow images are acquired by artificially masking some shadowless regions of the non-shadow natural images. However, The artificial shadowed images, would only cover limit shadow patterns, and visible differences between the actual shadowless images and their non-shadow labels would be induced, as have been demonstrated in \cite{hu19}.

\textbf{The unsupervised dataset}: 
Up to now, the USR \cite{hu19} dataset is the first and the unique unsupervised shadow removal dataset. It contains 2445 shadow images (1956 for training and 489 for testing) and 1770 non-shadow images. Note that the 1770 non-shadow images are not correlated with any image in the 2445 shadow images. Moreover, the USR dataset is captured on a thousands of different outdoor scenes and shaded by various types of objects, which covers the most number of different scenes and shadow patterns over other existing datasets.

\begin{table*}[t]
	\begin{center}
	\begin{tabular}{|l|c|c|c|c|c|}
		\hline
		 Methods &  Stage-I Rating&  Stage-II Rating &  FID \cite{FID2017} &  KID \cite{KID2018} &  RMSE \\
		\hline\hline
		 GAN-alone \cite{isola17} &  2.85$\pm$1.65&  2.31$\pm$ 1.96 &  108.4338 &  2.9889$\pm$ 0.4872 &  20.5989 \\
		 DistanceGAN \cite{benaim17} &  3.04$\pm$1.73&  2.67$\pm$2.18  &  103.8114 &  2.7214$\pm$ 0.4888 &  14.5482 \\
		 CycleGAN \cite{zhu17} &  4.36$\pm$2.02& 3.94$\pm$1.82  &  85.9442 &  2.4699$\pm$ 0.4858 &  12.7868 \\
		 U-GAT-IT [\cite{kim19} &  4.12$\pm$2.21& 3.99$\pm$1.58  &  95.6695 &  2.7124$\pm$ 0.4447 &  10.8764 \\
		 AttentionGAN \cite{tang19a} &  4.21$\pm$2.51& 3.28$\pm$2.78  &  95.1523 &  2.8461$\pm$ 0.4816 &  \textbf{4.3253}\\
		 Mask-ShadowGAN \cite{hu19} &  4.51$\pm$1.49& 3.46$\pm$1.57  &  85.0228 &  2.4096$\pm$ 0.4568 &  12.8541 \\
		 TC-GAN(ours) &  \textbf{5.91$\pm$1.35}& \textbf{6.66$\pm$1.62}  &  \textbf{72.3522} &  \textbf{1.6503$\pm$ 0.4548} &  10.5271 \\
		\hline
		
	\end{tabular}
	\end{center}
	\caption{ Qualitative and quantitative results on the USR dataset. For user study results (Stage-I and Stage-II ratings), higher is better. For FID, KID (KID$\times$100 $\pm$ std.$\times$100) and RMSE metrics, lower is better.}
	\label{table:tab1}
\end{table*}
\vspace{-1pt}
\subsection{Experiments on USR Dataset}
Since the difficulty and feasible data has not been available for a long time, the topic of unsupervised shadow removal has rarely been investigated. Mask-ShadowGAN \cite{hu19} is the state-of-the-art unpaired shadow removal methods and has shown to have good performance on the USR dataset. In addition to it, we compare our TC-GAN with some latest unsupervised image-to-image translation models, which are trained on the USR dataset by using the official provided source code and hyper-parameters. Specifically, the compared models are: GAN-alone (Pix2Pix \cite{isola17} model without paired supervision), DistanceGAN \cite{benaim17}, CycleGAN \cite{zhu17}, U-GAT-IT \cite{kim19} and AttentionGAN \cite{tang19a}. Noted that in order to make fair comparison, we generate shadow residual for all the generators as described in Section 3.1, and finally add it to the input shadow image to obtain the shadow-free output.

\vspace{-1pt}
\subsubsection{Qualitative Evaluation}
By following the qualitative evaluation on the USR dataset in Mask-ShadowGAN \cite{hu19}, we  conduct a user study through publishing a rating task on a source crowding platform to evaluate the shadow removal performance. However, in order to make further comparison on the visual results of the non-shadow area, we propose to conduct a two-stage subject experiment, which studies 10 participants in different ages and genders. In stage-I, each participant is given by randomly selected 10 shadow images from the USR testing set. We then apply all the methods to be compared to generate a total of 70 shadow-free images. Afterwards, we ask the participant to rate all the 70 shadow-free images in a scale from 1 (bad quality) to 10 (good quality). In Stage-II, we select another 10 shadow images randomly and perform the same settings as stage-I. Differently hereby, we provide participants with the original shadow images that correspond to the 70 shadow-free images for reference. In general, stage-I aims to evaluate the quality of the shadow-free images generated by different methods, while stage-II is to evaluate the visual effect of different shadow removal results through comparing from the input shadow images, i.e. check the consistency of non-shadow area between the recovered shadow-free image and original shadow image in terms of illumination, color and etc.

\begin{figure*}[t]
	\begin{center}
		\includegraphics[width=1.0\linewidth]{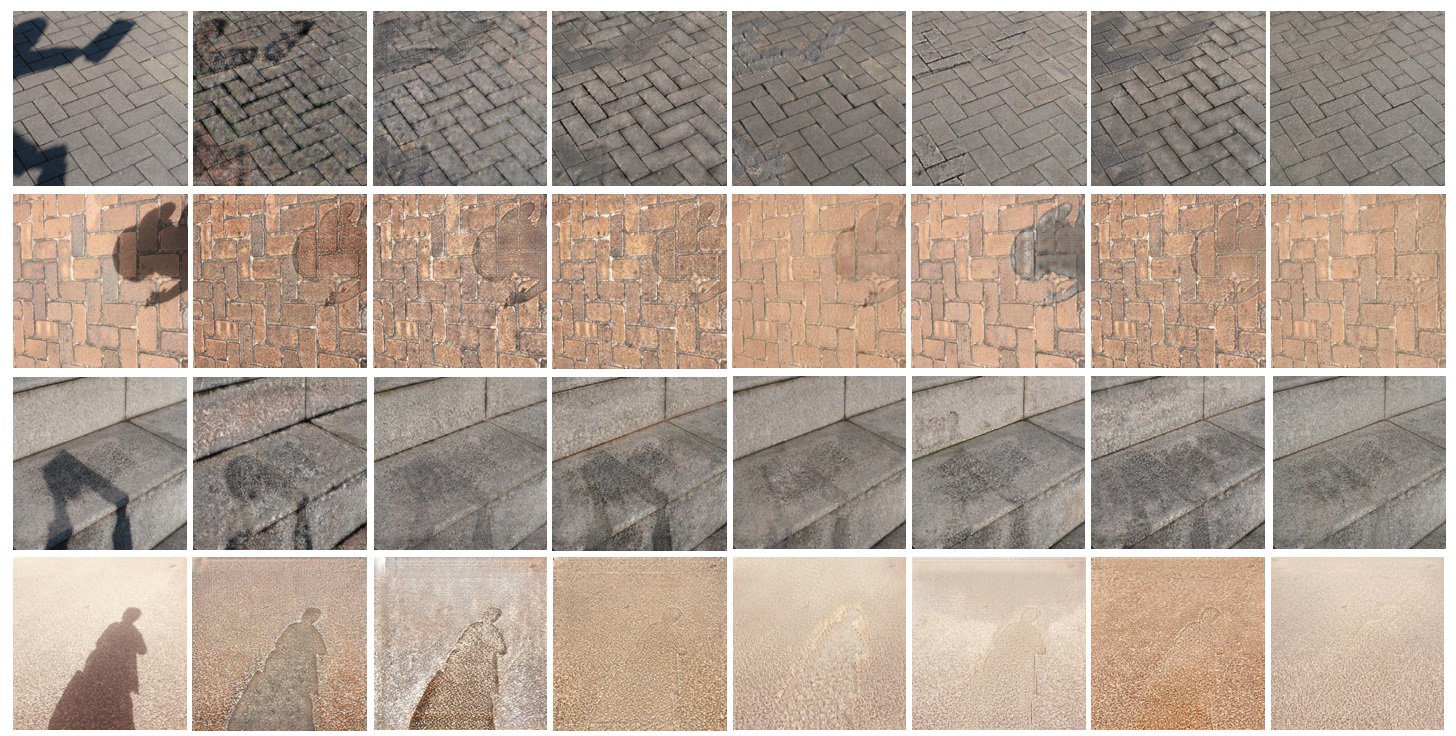}
		\put(-465,-8){\footnotesize (a)}
		\put(-405,-8){\footnotesize (b)}
		\put(-345,-8){\footnotesize (c)}
		\put(-285,-8){\footnotesize (d)}
		\put(-225,-8){\footnotesize (e)}
		\put(-160,-8){\footnotesize (f)}
		\put(-100,-8){\footnotesize (g)}
		\put(-40,-8){\footnotesize (h)}
	\end{center}
	\caption{Visual comparison of shadow removal results on USR dataset.
		(a) is input images.
		(b) is results of GAN-alone \cite{isola17}.
		(c) is results of DistanceGAN \cite{benaim17}.
		(d) is results of CycleGAN \cite{zhu17}.
		(e) is results of U-GAT-IT \cite{kim19}.
		(f) is results of AttentionGAN \cite{tang19a}.
		(g) is results of Mask-ShadowGAN \cite{hu19}.
		(h) is our TC-GAN's results.}
	\label{fig:fig4}
\end{figure*}
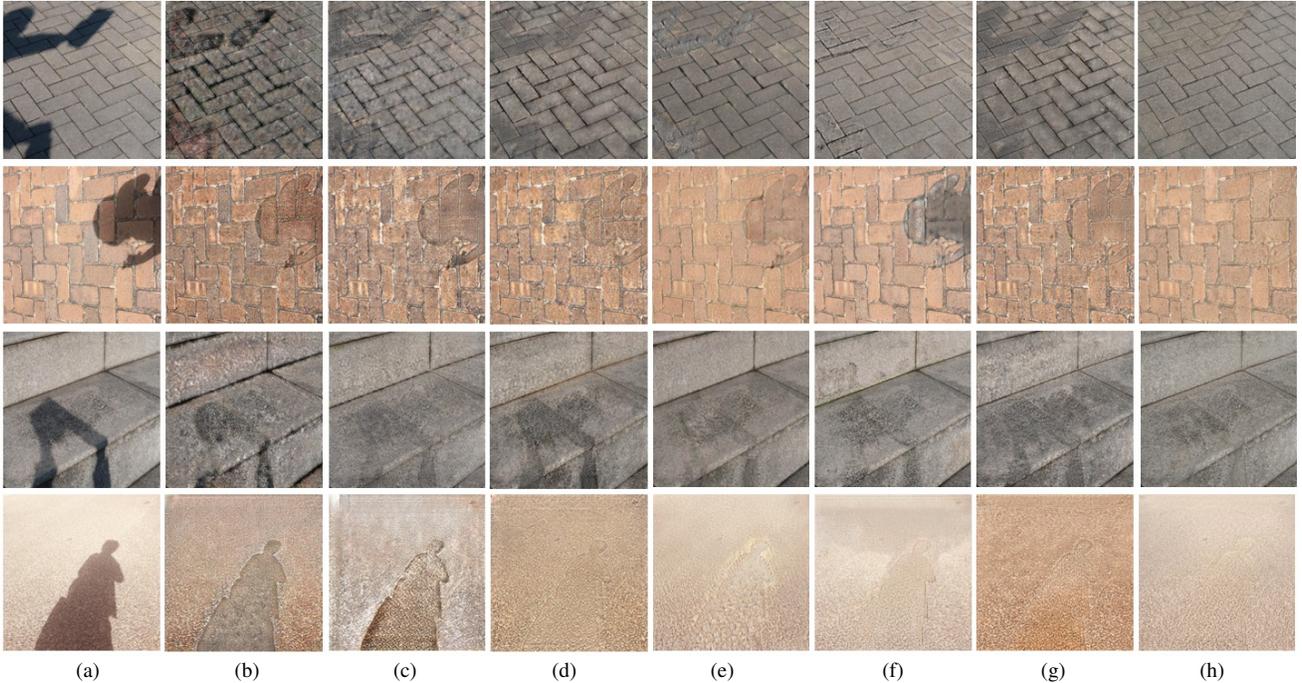
\vspace{-0.5pt}

In total, we obtain 100 ratings (10 participants × 10 images) per method for both Stage-I and Stage-II, respectively. We calculate the mean and standard deviation of these ratings and summarized the results in the second and third columns of Table \ref{table:tab1}, where the higher value of rating results indicating better performance. As can be seen that our TC-GAN achieves the highest rating performance in both Stage-I and Stage-II. Besides, although the cycle-consistency base models (CycleGAN, U-GAT-IT, AttentionGAN and Mask-ShadowGAN) achieve an improvement performance compared with vanilla GANs, they are not as good as our TC-GAN that is base on target consistency constraint. In addition, since the overall framework of TC-GAN is driven to maintain the correlation between the input shadow image and generated shadow-free image through learning, it also shows a greater advantage in the Stage-II ratings.

\vspace{-2pt}
\subsubsection{Quantitative Evaluation}
In order to make quantitative evaluation, which is not considered in Mask-ShadowGAN \cite{hu19}, we further calculate the FID \cite{FID2017} and KID \cite{KID2018} scores for all compared models. The FID and KID scores are two commonly used evaluation metrics in image-to-image translation field on the unpaired data, which aim at measuring the domain similarity between the real image and the transformed image domain. The lower of them indicate better generated results. Besides, in order to evaluate the correlation between the input shadow image and generated shadow-free target, we also measure the RMSE of the shadowless areas between them. Specifically, we use the BDRAR method \cite{zhu18} and the official pre-trained model to obtain the shadow mask for the shadow image. A smaller RMSE score represents a better match between shadow image and the output shadow-free image, which means the less variation is presented to the non-shadow regions.

The quantitative evaluation results of all the compared methods are shown in the last 3 columns of Table \ref{table:tab1}. Moreover, we present the visual comparison of shadow removal results of all methods on USR dataset in Figure \ref{fig:fig4}. As can be seen that compared with all other methods, our TC-GAN obtains the smallest FID and KID scores. It should be noted that since AttentionGAN directly copies the non-shadow pixels of the input shadow image to the output shadow-free image, it obtains a relative lower RMSE score. However, the worse results of AttentionGAN in terms of sujective rating, FID, KID and visual images indicate that AttentionGAN fails in removing shadow samples effectively.  In addition, it can be observed from Figure \ref{fig:fig4} that due to the absence of sufficient constraint, the generated image of GAN-alone and DistanceGAN contains many random perturbations and artifacts. Columns 4 to 6 in Figure \ref{fig:fig4} are the results of cycle-consistency based models. We observe that the output of these models retains some information of the shadow regions, leading to the failure of shadow removal. We believe the reason is that cycle-consistency constraint assumes that the generated shadow-free image could be reconstructed to the input shadow image without additional information. As a result, the generated shadow-free image has to maintain some shadow information. Mask-ShadowGAN facilitates reverse mapping by using the generated shadow mask. However, because the binary mask cannot provide intensity and color information of the shadow area, the shadow-free images generated by Mask-ShadowGAN have large appearance deviations from the real non-shadow ones. 

Comparatively, as shown in both Table \ref{table:tab1} and Figure \ref{fig:fig4}, the proposed TC-GAN not only achieves the best qualitative and quantitative scores, but is able to completely remove shadow regions while keeping the high correlation of the non-shadow area between the recovered shadow-free image and the input image.

\vspace{-1pt}
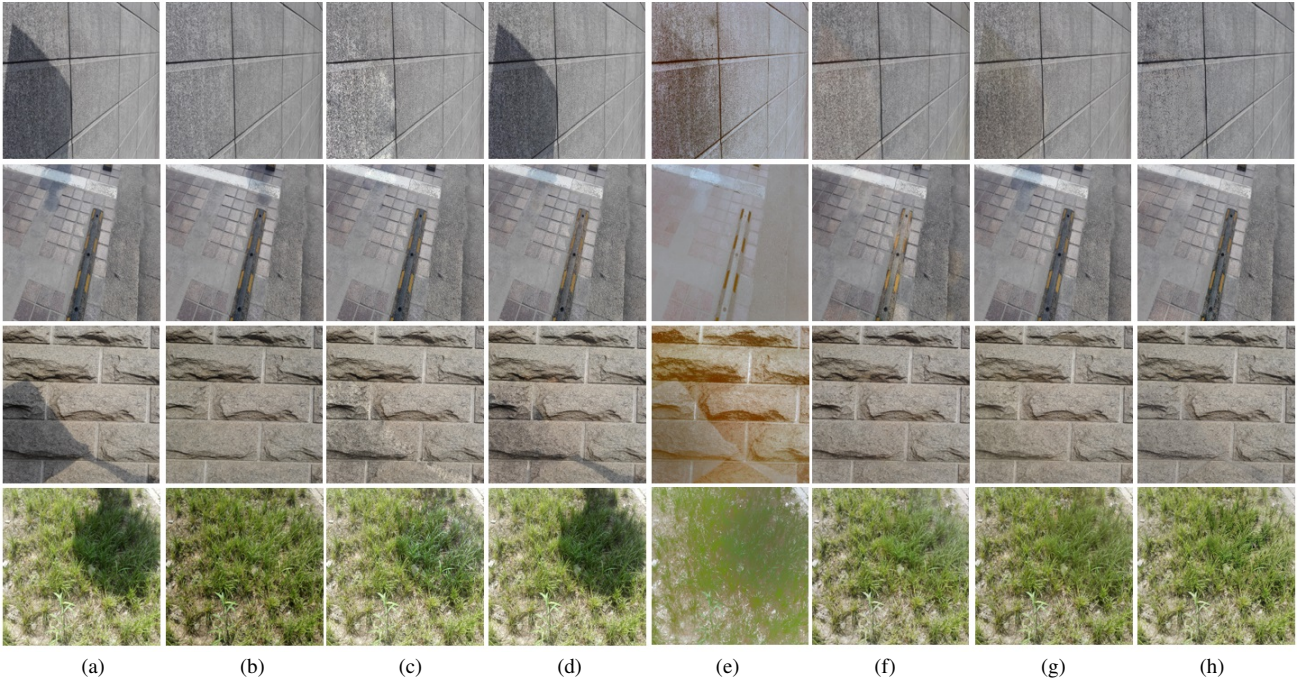
\begin{figure*}[t]
	\begin{center}
		\includegraphics[width=1.0\linewidth]{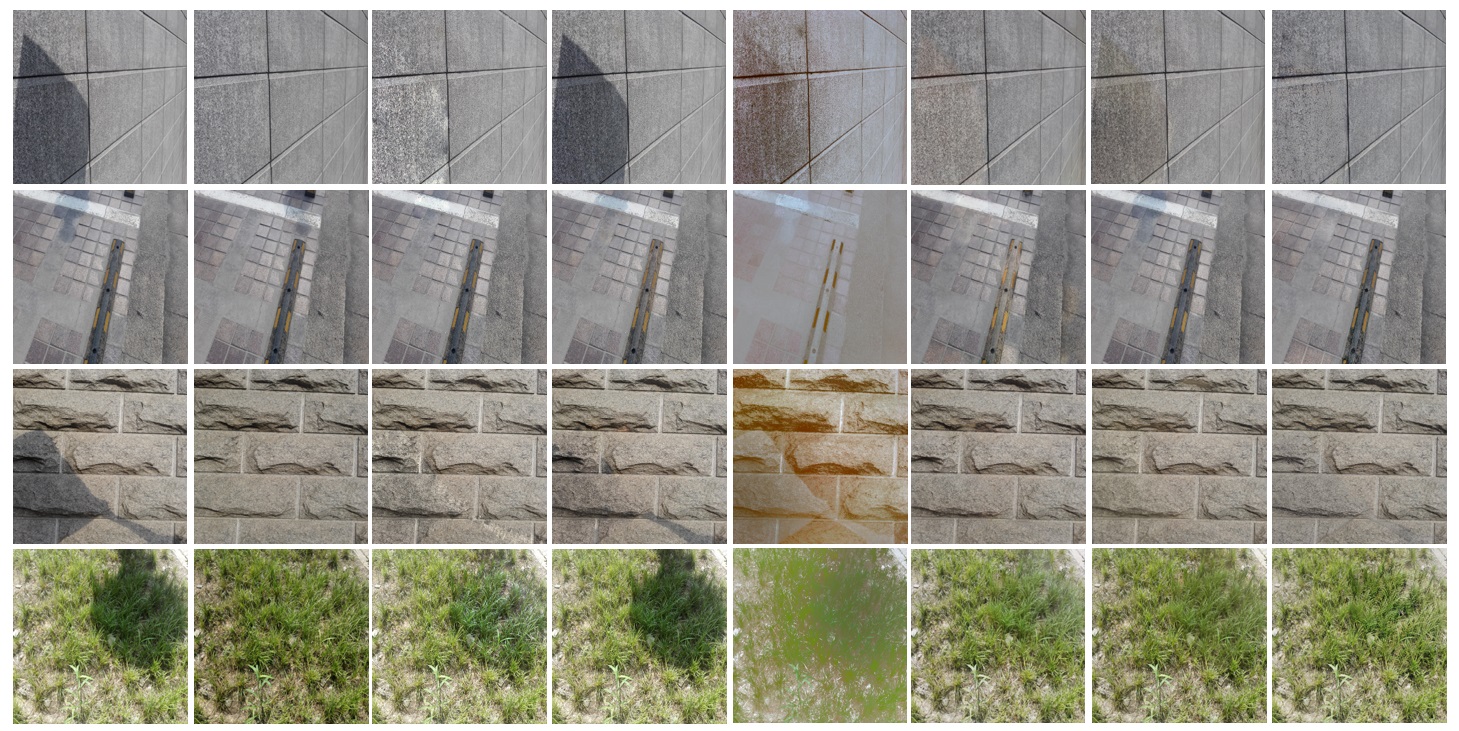}
		\put(-463,-8){\footnotesize (a)}
		\put(-403,-8){\footnotesize (b)}
		\put(-343,-8){\footnotesize (c)}
		\put(-283,-8){\footnotesize (d)}
		\put(-223,-8){\footnotesize (e)}
		\put(-163,-8){\footnotesize (f)}
		\put(-100,-8){\footnotesize (g)}
		\put(-40,-8){\footnotesize (h)}
	\end{center}
	\caption{Visual comparison of shadow removal results on ISTD dataset.
		(a) is input images.
		(b) is ground truth images.
		(c) is results of Gong \etal \cite{gong14}.
		(d) is results of Guo \etal \cite{Guo12}.
		(e) is results of Yang \etal \cite{yang12}.
		(f) is results of ST-CGAN \cite{wang18}.
		(g) is results of DSC \cite{hu18}.
		(h) is our TC-GAN's results.}
	\label{fig:fig5}
\end{figure*}
\vspace{-2pt}

\subsection{Experiment on ISTD Dataset}

In order to further evaluate our methods, we compare our TC-GAN with some supervised methods on the supervised ISTD dataset as well. Here, many baseline and advanced shadow removal methods  are take into comparison on the ISTD dataset, including Gong \etal \cite{gong14}, Guo \etal \cite{Guo12}, Yang \etal \cite{yang12}, ST-CGAN \etal \cite{wang18} and DSC \cite{hu18}. Specially, we obtain their results and source code directly from the official website and in order to make a fair comparison, we re-scale the size of all the testing images to 256 × 256 via cubic spline interpolation.  In addition, since TC-GAN is independent of shadow mask information, which is provided by the ISTD dataset, while many supervised have used it as the guidance to recover the shadow-free image, we do not compared with more state-of-the-art methods that use full ISTD for training. We evaluate the performance of all the supervised methods by calculating RMSE between the generated shadow-free image and the annotated ground truth by considering several validation areas: (i) in shadow area (S), (ii) non-shadow area (N) , (iii) the entire image (A) respectively. In addition, by considering that the non-shadow labels in the ISTD dataset contain noises, we further measure the RMSE of the generated shadow-free image and the input shadow image in the non-shadow area (N-I).
\vspace{-2pt}
\begin{table}[h]
	\begin{center}
	\begin{tabular}{|l|c|c|c|c|}
		\hline
		 Method & S & N & A & N-I \\
		\hline\hline
		 Gong \etal \cite{gong14} & 14.98  & 7.29  & 8.53  & -  \\
		 Guo \etal \cite{Guo12} & 18.95  & 7.46  & 9.3  & - \\
		 Yang \etal \cite{yang12} &18.92  & 14.83 & 15.63  & - \\
		 ST-CGAN \cite{wang18} & 10.33 & 6.93 & 7.43 & 7.45 \\
		 DSC \cite{hu18} & \textbf{9.22}  & 6.39  & \textbf{6.67}  & 6.61 \\
		 TC-GAN & 11.49 & \textbf{5.91} & 6.85 & \textbf{6.29}\\
		\hline
	\end{tabular}
	\end{center}
	\caption{RMSE results for different methods on ISTD dataset. Abbreviations: S (shadow area), N (non-shadow area), A (entire image) and N-I (compared with non-shadow area of input image).}
	\label{table:tab2}
\end{table}
\vspace{-5pt}
For our TC-GAN on the ISTD dataset, we disrupt the paired order of shadow images and non-shadow images during training, and use random cropping to avoid pixel matching between any two shadow images and non-shadow image in the dataset. We finally summarize all the evaluation results  Table \ref{table:tab2}. As can be seen that our TC-GAN achieves comparable performance over the supervised methods, especially for the non-shadow area (N-I) results, and importantly, TC-GAN is trained without paired supervision. 

Figure \ref{fig:fig5} illustrates the visual comparison of shadow removal results on ISTD dataset. It can be seen that due to the noise of training labels, the supervised methods usually cannot remove shadow perfectly (the first and second rows) or keep the non-shadow areas unchanged (the third and fourth rows). Comparatively, TC-GAN effectively removes the shadows and retain the content and illumination in non-shadow areas simultaneously, even if the paired data are not used for training.

\vspace{-2pt}
\section{Conclusion}
In this work, we propose a novel target consistency generative adversarial network, named TC-GAN, for unsupervised shadow removal. Our TC-GAN uses target consistency constraint to learn a unidirectional mapping from shadow to non-free domain. The overall network is designed to be a dual GAN structure. The correlation between the generated shadow-free target and input shadow image and the authentication of the generated shadow-free image is strictly confined through the whole network training. The proposed TC-GAN is compared with many state-of-the-art unsupervised shadow removal and supervised methods in terms of both quantitative and qualitative evaluation, results demonstrate the superior performance and the good recovered quality of our method. In the future, we plan to apply TC-GAN to more shadow scene tasks, such as shadow detection and shadow removal in video, etc.

{\small
\bibliographystyle{ieee_fullname}
\bibliography{egbib}
}

\clearpage
\setcounter{table}{0}
\setcounter{figure}{0}
\setcounter{section}{0}

\section{Discussion about Shadow Transform Encoders (STE)}
In the proposed target-consistency generative adversarial network (TC-GAN), the two shadow transform encoders (STE) in the two corresponding generators attempt to encode the same shadow image into different feature combinations so as to satisfy the target consistency assumption for the shadow removal problem. In practice, we do not explicitly constraint the feature combinations of the two shadow transform encoders to be different, as this is relatively complex and not conductive to convergence stably during training, but provide different adversarial optimization objectives for the two generators at each iteration during training by feeding with different real non-shadow images for the two discriminators respectively. In order to verify that the STE has learned different feature combinations from the input shadow image separately, we visualize some feature maps produced by the two STE with respect to the same shadow domain input in Figure \ref{fig:fig2}. Note that the full feature maps of each STE contains 256 channels, we present the activation maps from the first 10 channels for the best look. As we can see from Figure \ref{fig:fig2}, the feature maps obtained by the two STE are significantly different in activated information for each channel, and as is expected that the input shadow image has been transformed into different feature combinations of the shadow or non-shadow semantic objects. During the overall training process, although with different generated features and decoded shadow residuals by the shadow residual decoders, the two consistent shadow-free images that are corresponding to the same shadow input can be generated by the jointly optimization of adversarial generative constraint, the target consistency constraint and the identity loss. 

\begin{figure*}[!t]
	\begin{center}
		\includegraphics[width=1.0\linewidth]{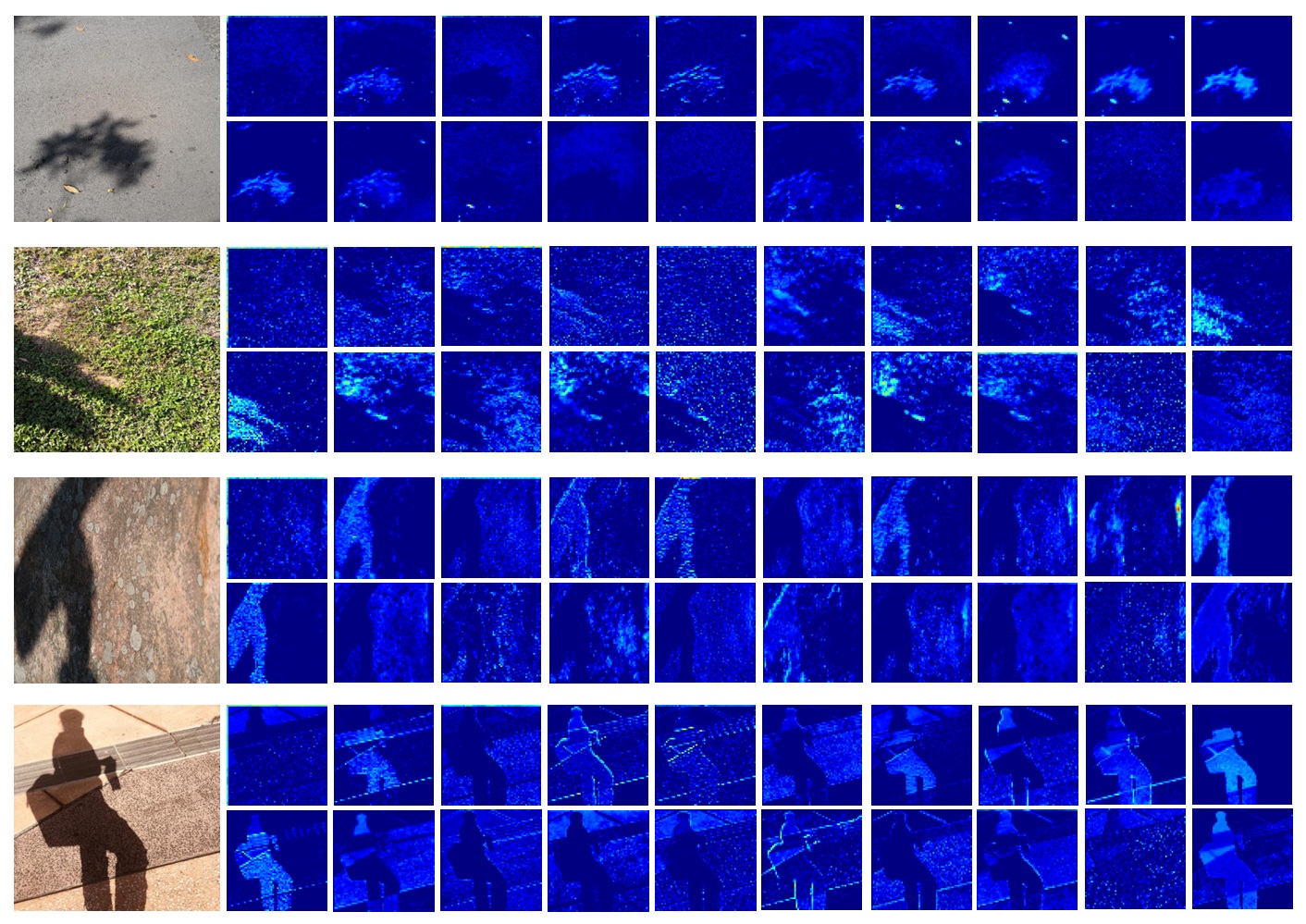}
	\end{center}
	\caption{Visualization of the output heatmaps of different shadow transform encoders (STE). The first column is the input shadow images, followed by the feature heatmaps generated by STE, where the top and bottom are the first 10 channels of feature heatmaps output by two STE respectively.}
	\vspace{-0.5pt}
	\label{fig:fig2}
\end{figure*}

\begin{table}[b]
	\begin{center}
		\begin{tabular}{c|c|c|c}
			\hline
			\hline
			Selection Strategy & $G_1$ & $G_2$ & MSM \\ \hline
			FID & 75.1297 & 74.9462 & \textbf{72.3522} \\ 
			\hline
			\hline
		\end{tabular}
	\end{center}
	\caption{FID scores for different model selection strategy on USR dataset, and for this metric, lower is better. $G_1$ and $G_2$ are two mapping networks and MSM is our model selection module.}
	\label{table:tab1}
\end{table}

\section{Discussion about Module Selection Module (MSM)}

Our TC-GAN contains two completely symmetric mapping networks from the shadow domain to non-shadow domain. In theory, the two mapping networks should perform almost identically on the shadow removal task, but for each specific sample, the output of two different networks may differ slightly. Therefore, in the testing phrase, we need to choose one of the output shadow-free images from the two mapping networks as the final shadow-free target of TC-GAN, which is the role of the model selection module (MSM). In fact, different purposes usually correspond to different model selection strategies. For example, for a hardware and time sensitive application, the output of one of the mapping network can be fixed directly as the final output, and the other mapping network can be discarded, which can reduce the number of parameters and inferring time by half. In this work, we propose a simple and reasonable model selection strategy. We add a shadow/non-shadow binary classifier that is pre-trained with the shadow removal dataset to the model selection module, and the output scalar of the classifier represents the probability that the input shadow-free image is a real non-shadow image. The network structure of the model selection classifier is the same as that of the discriminator, which makes it have enough classification ability. The classifier is trained in advance, and finally reached the correct rate of 90\% on the testing dataset. Therefore, during the testing phrase, for each sample, we choose the output of two mapping networks with a higher probability of real non-shadow on the model selection classifier as the final output of TC-GAN. In the experiments, we found that the performance of using model selection module improve slightly compared to the performance of directly fixing one of the mapping networks. Specifically, taking the FID evaluation metric in the USR unpaired dataset as an example, as shown in Table \ref{table:tab1} the optimal FID scores of the two mapping networks are 75.1297 and 74.9462 respectively, while the score of the use of model selection module is 72.3522, whose performance is improved by 3.46 \%.

\begin{figure*}[t]
	\begin{center}
		\includegraphics[width=1.0\linewidth]{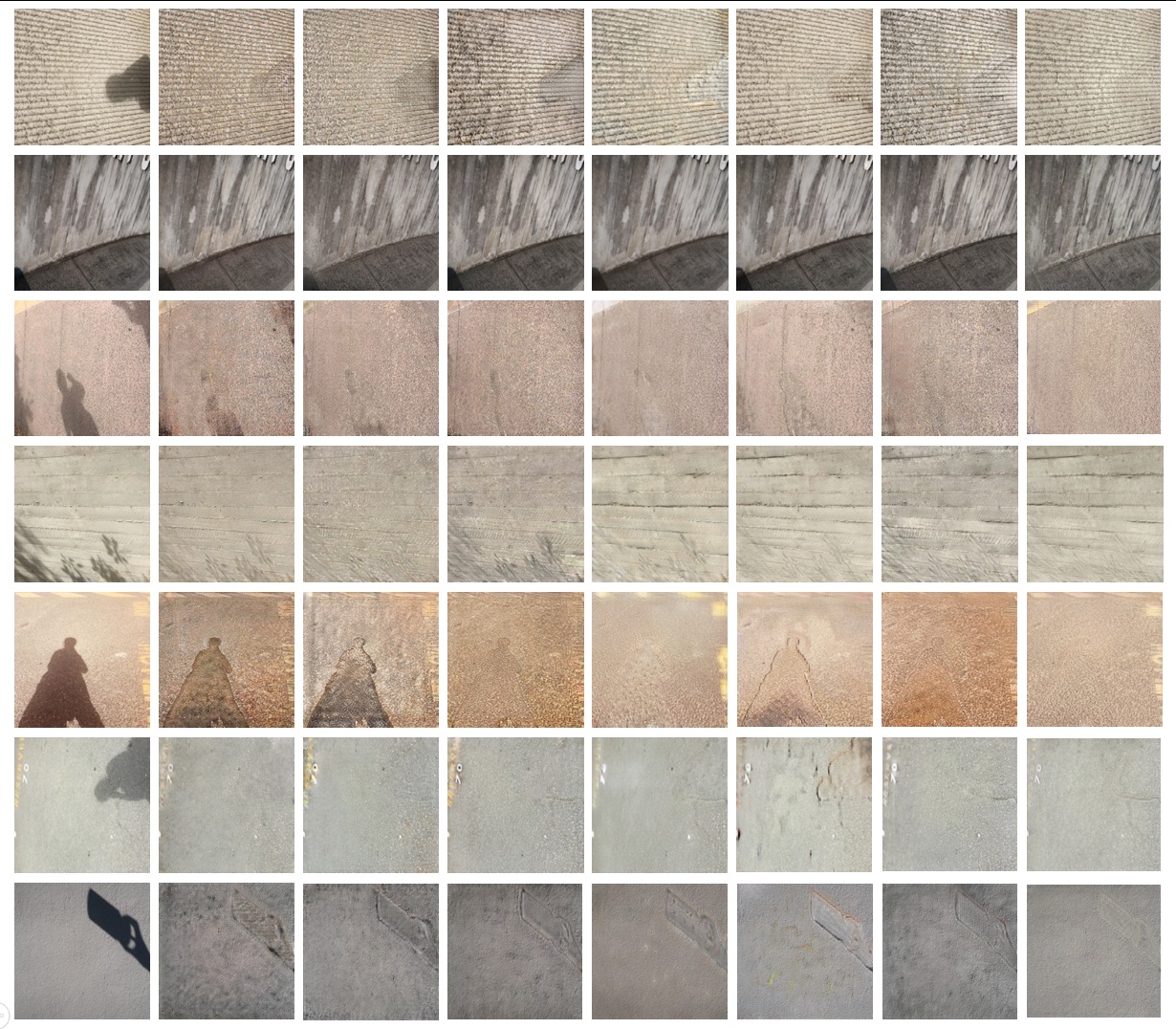}
		\put(-465,-8){\footnotesize (a)}
		\put(-405,-8){\footnotesize (b)}
		\put(-345,-8){\footnotesize (c)}
		\put(-285,-8){\footnotesize (d)}
		\put(-225,-8){\footnotesize (e)}
		\put(-160,-8){\footnotesize (f)}
		\put(-100,-8){\footnotesize (g)}
		\put(-40,-8){\footnotesize (h)}
	\end{center}
	\caption{More visual comparison of shadow removal results on USR dataset.
		(a) is input images.
		(b) is results of GAN-alone.
		(c) is results of DistanceGAN.
		(d) is results of CycleGAN.
		(e) is results of U-GAT-IT.
		(f) is results of AttentionGAN.
		(g) is results of Mask-ShadowGAN.
		(h) is our TC-GAN's  results.}
	\vspace{-0.5pt}
	\label{fig:fig3}
\end{figure*}

\begin{figure*}[t]
	\begin{center}
		\includegraphics[width=0.7\linewidth]{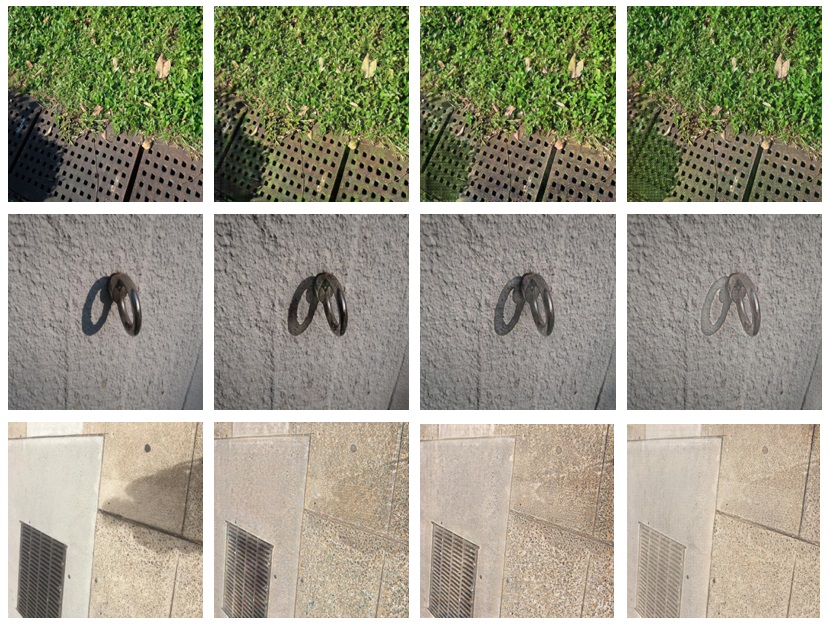}
		\put(-210,-8){\footnotesize (a)}
		\put(-150,-8){\footnotesize (b)}
		\put(-90,-8){\footnotesize (c)}
		\put(-30,-8){\footnotesize (d)}
	\end{center}
	\caption{Some examples of failure for shadow removal. (a) is input shadow images, (b) is results of CycleGAN, (c) is results of Mask-ShadowGAN and (d) is results of proposed TC-GAN.}
	\vspace{-0.5pt}
	\label{fig:fig1}
\end{figure*}

\section{More Examples of Visual Comparisons}

In Figure \ref{fig:fig3}, we provide more examples of visual comparisons on the USR dataset. It can be seen from the figure that our TC-GAN can achieve better performance than other unsupervised image-to-image translation and unpaired shadow removal approaches, both in terms of the effect of shadow removal and pixel retention in non-shadow regions. However, due to the lack of supervision between the shadow image and the corresponding non-shadow label, TC-GAN may fail to remove shadows from the shadow image with some complex scenes. Three typical scenarios are shown in Figure \ref{fig:fig1}. TC-GAN may restore the shadow to the wrong color when the shadow area covers a variety of backgrounds and large color difference. As shown in the first row of Figure \ref{fig:fig1}, TC-GAN restore the shadow region to green mistakenly because it mistaken the background of the shadow region for a lawn. When the shadows in the image are very similar to the object in the non-shadow area, TC-GAN may change the non-shadow region, as shown in the second row of Figure \ref{fig:fig1}. In addition, when there are objects in the non-shadow region of the shadow image that are darker than the shadows, TC-GAN is likely to remove them simultaneously. As shown in the last row of Figure \ref{fig:fig1}, TC-GAN treats the manhole cover as an image shadow mistakenly and tries to remove it. In general, we find that when dealing with these complex scenes, other methods are either unable to remove the shadow of the image or suffer from the same problem as our TC-GAN, which is also the direction of the unsupervised shadow removal methods to be optimized in the next step.

\end{document}